\documentclass[letterpaper, 10pt, conference]{ieeeconf}
\IEEEoverridecommandlockouts

\usepackage{graphicx}
\usepackage{adjustbox}
\usepackage{multirow}
\usepackage{amssymb}
\usepackage{color}
\usepackage{tikz}

\usepackage{lineno}
\usepackage{soul}
\usepackage{tabularx, rotating}
\usepackage{array}
\usepackage{siunitx}
\usepackage{url}
\usepackage{booktabs}
\usepackage{comment}
\usepackage{mathtools}
\usepackage{pifont}
\usepackage{cite}
\usepackage[normalem]{ulem}

\usepackage{subcaption}
\usepackage{verbatim}

\newcommand \red[1]{\textcolor{red}{#1}}

\usepackage[hidelinks]{hyperref}

\usetikzlibrary{external}
\usepackage[T1]{fontenc}
\usepackage[utf8]{inputenc}
\usetikzlibrary{positioning}
\usetikzlibrary{angles, quotes}

\graphicspath{ {./images/} }
\begin{document}



\title{\LARGE \bf Frontier-Based Exploration for Multi-Robot Rendezvous in Communication-Restricted Unknown Environments}

\author{Mauro Tellaroli, Matteo Luperto, Michele Antonazzi, Nicola Basilico
\thanks{All authors are with the Department of Computer Science, University of Milan, Milano, Italy {\tt\small~name.surname@unimi.it}
}
}

\maketitle

\begin{abstract}

Multi-robot rendezvous and exploration are fundamental challenges in the domain of mobile robotic systems. This paper addresses multi-robot rendezvous within an initially unknown environment where communication is only possible after the rendezvous. Traditionally, exploration has been focused on rapidly mapping the environment, often leading to suboptimal rendezvous performance in later stages.
We adapt a standard frontier-based exploration technique to integrate exploration and rendezvous into a unified strategy, with a mechanism that allows robots to re-visit previously explored regions thus enhancing rendezvous opportunities.
We validate our approach in  3D realistic simulations using ROS, showcasing its effectiveness in achieving faster rendezvous times compared to exploration strategies.
\end{abstract}


\section{Introduction}

Multi-robot rendezvous involves coordinating a system of mobile robots, or a Multi-Robot System (MRS), to efficiently converge at or near a shared location. A rendezvous strategy is typically evaluated by the total time or distance taken for all robots to reach the meeting point, the smaller the better. These metrics are often interpreted as proxies for both energy consumption and task efficiency.

Computing and executing efficient rendezvous strategies represent key components in MRS application domains where robots need to physically meet to share collected information or collaborate on some localized task. Distributed data-gathering offers many settings where multi-robot rendezvous is required or can play a fundamental role. Examples include autonomous exploration~\cite{corah2017efficient}, persistent surveillance or monitoring~\cite{iocchi2011multi,liu2021team}, and search and rescue~\cite{liu2016multirobot}. 
The distributed nature of these tasks reflects the lack of a centralized infrastructure covering the whole environment and enabling global communication and coordination among robots. Instead, robots have to rely on peer-to-peer interactions, which, being subject to minimum-range constraints, require physical proximity. Meeting with teammates enables sharing partial maps in exploration or exchanging findings and data collected in monitoring and search, hence allowing robots to compute more informed plans for solving a common task. Communication is not the only domain where rendezvous might play a role. Multi-robot task allocation scenarios~\cite{khamis2015multi} often feature tasks that, to be executed, require concurrent cooperation by more robots (perhaps also heterogeneous ones); in such cases, meeting at the same (task) location is a pre-condition. 

\begin{figure}[t]
    \centering
    \includegraphics[width=\linewidth]{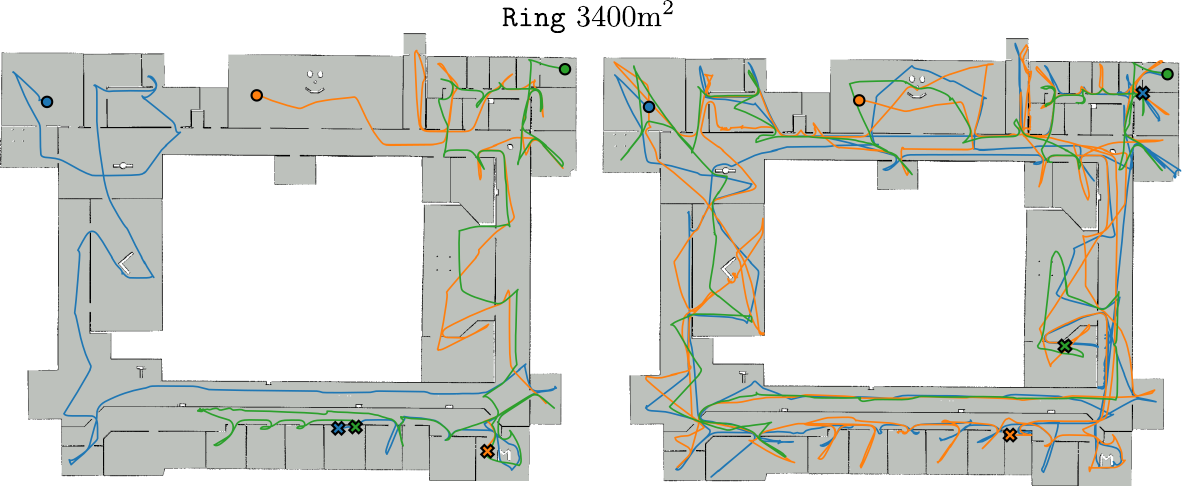}
    
    \caption{Rendezvous paths of $3$ robots in an initially unknown, communication-restricted environment, using our method (Frontier-Based Rendezvous, \texttt{FBR}, left) and the classical frontier exploration strategy (Frontier-Based Exploration, \texttt{FBE}, right) in the map \texttt{Ring}, $3400 \text{m}^2$.
     Circles indicate starting positions, while crosses indicate the locations when a robot joins a cluster. When robots are in the same cluster, we indicate only the trajectory of the leader. Each robot has its own map and its own SLAM module; we use the full map of the environment for visualization purposes.}
     \vspace{-.5cm}
    \label{fig:path_example}
\end{figure}

In this work, we consider the rendezvous problem for a team of autonomous mobile robots in the challenging setting of a communication-restricted~\cite{banfisurvey} and initially unknown indoor environment. We assume that the robots start from different arbitrary locations, but no map of the environment is available to any of them and no pre-determined meeting location or coordination strategy has been agreed upon. 
The communication in the environment is restricted to occur only after a rendezvous.
An example of this problem is when two or more people need to gather in an environment they don't know, such as in a shopping mall, a hospital, or an airport. 

In this scenario, the difficulty of the rendezvous problem is augmented by the fact that the MRS needs to simultaneously perform an online exploration task, as the environment is unknown. Exploration is the problem of navigating an initially unknown environment to build its accurate map, and it is customarily performed by iteratively (i) identifying a set of promising candidate locations in the already mapped portion of the environment,  (ii) selecting the most promising one according to a set of criteria, and  (iii)~reaching the target location while integrating new perceptions into the partial map of the environment.
A popular choice in step~(i) is to use \textit{frontiers}~\cite{Yamauchi98}, namely the boundaries between the explored and unexplored part of the environment, to determine candidate locations to reach, and to select the most promising one according to an \textit{exploration strategy}. When an MRS is involved, each robot can carry out its exploration independently, or they can coordinate by communicating and exchanging maps and plans~\cite{rogers2013coordination}.

Frontier-based exploration strategies are widely used as they are simple yet effective, allowing robust exploration in heterogeneous contexts. However, the inherent greedy nature of frontier-based exploration (see (i)-(iii) above), which aims to obtain the largest map of the environment in the shortest time, results in quickly mapping those regions where the most free space lies, leaving behind frontiers in peripherical and less informative portions of the environment, that are eventually explored in later stages of the exploration run \cite{RAS2020}. Such a rationale might be in contrast with what is required by an efficient rendezvous, which, in principle, does not strictly require building a complete map.

Motivated by real-world concrete mobile-robot settings, we propose a definition of rendezvous that enforces connectivity (to ensure communication) and promotes (without enforcing) physical proximity. Such a practical definition departs from the classical one typically studied in theoretical frameworks, where exploration and rendezvous are addressed on discrete representations like graphs or 2D polygonal environments (neglecting perceptions and navigation errors). The method we propose is an exploration strategy that is biased towards rendezvous, where frontiers are created not only for their potential contribution to the map's expansion (classical frontier-based approach) but also to increase the opportunity of meeting a teammate.
To do so, we introduce a mechanism that allows each robot to \textit{forget} about parts of the environment that have been explored and mapped, thus putting those parts back into the portions of the environment still to be explored. In this way, the robot is encouraged to backtrack on previously explored areas, often passing through high-connectivity zones as corridors, facilitating accidental rendezvous.
We evaluate this method in ROS with 3D realistic simulations of complex large-scale environments. The results obtained from our extensive experimental campaign show how the proposed can extend existing frontier-based exploration frameworks to achieve rendezvous with different teams of robots efficiently.

A first version of this work has been presented in \cite{AIRO} where we provided an overview of our method and a preliminary experimental campaign. This work provides an extension where we present the method in detail and conduct an extensive experimental campaign considering additional configurations and evaluation metrics. 

\section{State of the art}
Methods for multi-robot rendezvous typically leverage pre-determined coordination strategies among robots (pre-established rendezvous areas, biases applied to strategies to search for others, etc.) or execute online search strategies in a commonly known map to meet others. One distinction among existing approaches can be made between symmetric and asymmetric rendezvous. In symmetric rendezvous, all robots have the same role in seeking a meeting with others. In the asymmetric case, instead, some robots can be explorers, while others are relays, i.e., they have to meet and transfer knowledge acquired by other robots to each other or to a base station. 
Another distinction is between intentional rendezvous and accidental rendezvous. The former ones happen when the meeting is already scheduled among agents following some kind of coordination. In the latter ones the robots know that they need to meet but no time and place is pre-arranged. Other approaches undertake an offline study of the problem, by deriving theoretical properties of the optimal solution from the specific geometrical features of the environment~\cite{ozsoyeller2019rendezvous}.
Several of these approaches are not directly applicable to our scenario, where the target environment and starting locations of the MRS are unknown, and communication is restricted.

Multi-robot rendezvous and exploration have been studied together, but are often dealt with as mutually-exclusive alternating phases: when the robots are exploring, their interest is to acquire new knowledge; when the robots decide to rendezvous, they travel to a location in the mapped area.
The decoupling of the two phases can lead to suboptimal performance, as robots must switch between behaviors in a coordinated way.
This issue is investigated in~\cite{hourani2013serendipity} by introducing the concept of serendipity in exploration, i.e., to create a robot behavior that tries to facilitate an unplanned encounter with other robots while those are still in an exploration phase, by adopting a behavior that is a mix between episodic and planned rendezvous.

The work of~\cite{spirin2014rendezvous} investigates how to perform arranged rendezvous in an asymmetric system where a robot is a relay of information, the others are \emph{explorers}, and a rendezvous is set up in locations that may not be directly connected but are in communication range (e.g., are separated by a wall).

In~\cite{andries2013multi} the robots, after performing a full exploration, do a rendezvous in a location of the final map of each robot, using an ant-based strategy. 

The work of~\cite{6386049} presents a framework for the simultaneous multi-robot exploration and rendezvous on graphs. The MRS, without any communication, decides where to perform a rendezvous using three different strategies, namely asymmetric, symmetric, and exponential. In the asymmetric strategy, one of the robots is stationary at a node while the others explore the environment. The roles of the robots are pre-determined. In the symmetric strategy, all robots are exploring the environment.
In the exponential strategy, the MRS uses an exponentially decreasing function to estimate rendezvous locations.
At the beginning, the robots perform exploration using a breath-first strategy. After a time budget is reached, the robots try to attempt a rendezvous by going to high-likelihood rendezvous locations.
This approach is used also in~\cite{5957545} to minimize the rendezvous time while having a maximum speed gain of the exploration task, with the least possible dependency on communication. A rendezvous strategy is used to rank and visit locations promising to attempt a rendezvous. 

The work of~\cite{dah2023search} presents a framework similar to ours where to perform exploration, search, and rendezvous in a real-world deployment with line-of-sight communication, by relying on the content of virtual stigmergy~\cite{pinciroli2016buzz}. However, the goal of such a work is significantly different from ours, as it aims to perform periodic rendezvous at a pre-arranged time to share the information acquired during single-robot search tasks. After that, the MRS sets up a relay tree to connect the team to a base, sharing the position of the targets identified.


In~\cite{pittol2022loop}, similarly to our work, new sets of frontiers are added to those used by a frontier-based exploration method to facilitate loop closures, that improve map quality. In our work, we go beyond hallucinating frontiers to facilitate rendezvous.
\section{Our Method}

\subsection{Problem Formulation}

We consider a team of $m$ homogeneous robots equipped with the same perception, navigation, and communication capabilities. We assume a strongly limited communication model: each robot has a communication range of radius $d$ and, to exchange any information, two robots must be inside each other's range and have an unobstructed line of sight. Each robot starts from a random location, unknown to the others, in a given environment whose map is initially unknown. Our objective is to have the robots form a connected group so that, from that moment on, they can collectively plan and navigate in the environment while keeping a formation.

Formally, we define the rendezvous condition by requiring that the $m$ locations occupied by the team members satisfy a hard and a soft constraint. The hard constraint requires the robots to form a connected configuration as per the communication model described above (so each robot should, in principle, be able to communicate with another one in a multi-hop fashion). The soft constraint requires that robots should be close to each other to navigate in a formation. Ideally, this requirement should enforce a maximum distance between the locations of any pair of robots. However, setting it as a hard constraint to reach a rendezvous would in many cases prevent the problem's feasibility. Indeed, with many robots and environments that do not offer open-area regions, connected groups of robots might not find a way to maneuver into a joint configuration that, without collisions, realizes the required mutual uniform-distance formation. For this reason, in our method, we seek this condition as much as possible while allowing for its violation when no other option is available.


\subsection{Adapting Frontier-Based Exploration for Rendezvous}

Our proposed method is based on extending the frontier-based exploration approach~\cite{Yamauchi98}, to incentivize a fast episodic rendezvous. In the traditional single-robot scenario, exploration is carried out by iteratively integrating the robot's perceptions into an incremental 2D map of the environment. This map represents all those areas that the robot has explored (sensed) up to the present time. Within these maps, \emph{frontiers} are defined as the edges between parts of the map that have been explored and are obstacle-free, and those regions of the environment that are still unknown. We extend this setting to a multi-robot context, equipping robots with a frontier-based exploration strategy that they execute both independently and in an asynchronous manner. In each robot, the exploration process unfolds by iteratively following a sequence of steps:
\begin{enumerate}
\item from the current map, extract the set of frontiers $F$; from each frontier $f \in F$, compute a candidate location to reach. In our case this will be its centroid;
\item rank frontiers according to one or more criteria (e.g., the distance from the robot) aimed at determining the effectiveness of continuing the exploration by traveling there, and selecting the best one;
\item plan and execute a path to reach the centroid of the selected frontier, integrating into the map the perceptions acquired while traveling;
\item repeat from (1) until $F$ results empty.
\end{enumerate}

Using this framework, the robot aims to reach the boundaries of its current map, ideally without going back on its steps but continuously moving forward, until the whole environment is mapped.
In our method, we bias exploration by introducing an \textit{information decay} mechanism on the mapping process so that the robot is also driven to go back on its steps, following the intuition that this backtracking mechanism will promote episodic rendezvous among robots.

\begin{figure}
    \centering
    \includegraphics[width=.8\linewidth]{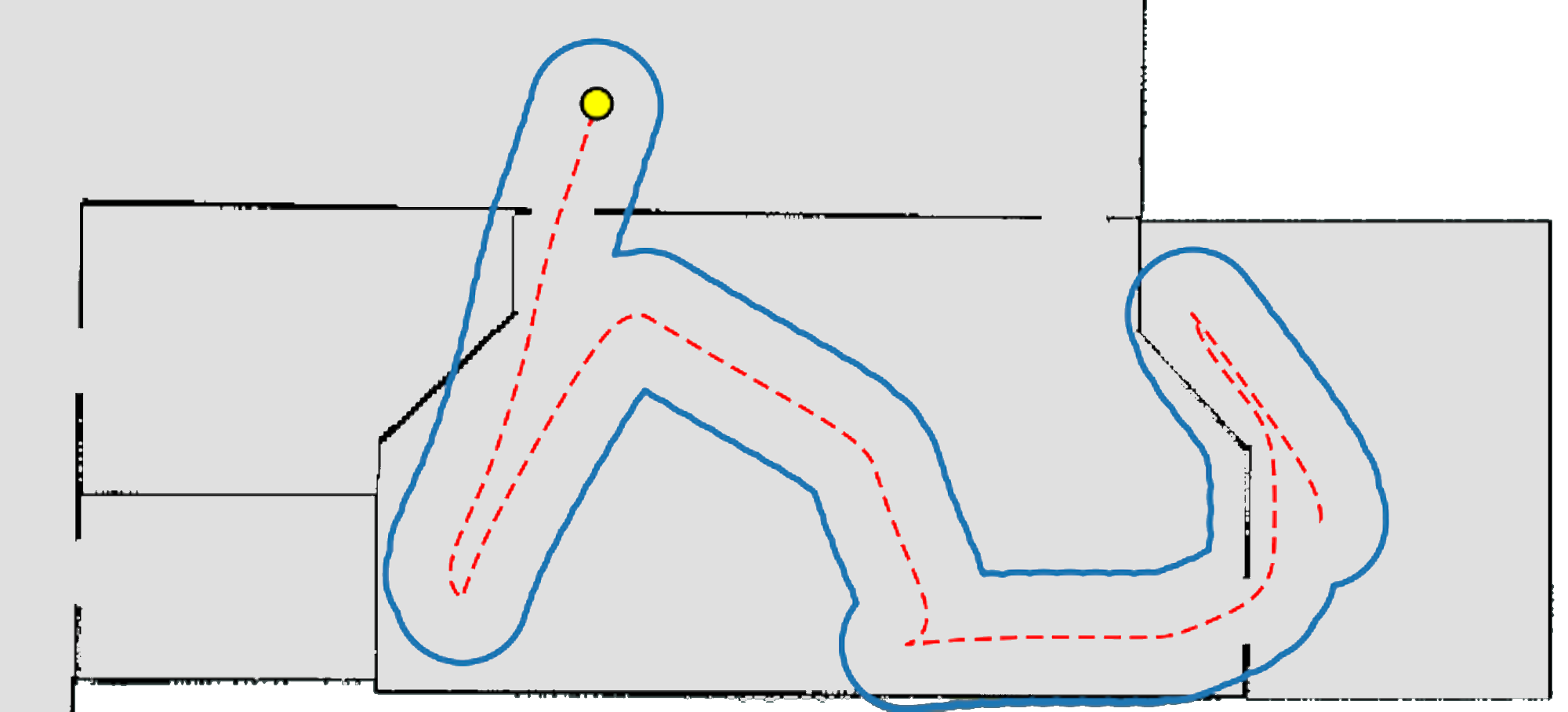}
    \caption{The  \textit{exploration trace}, circled in blue, of the robot that followed the trajectory highlighted in dashed red.}
    \label{fig:blob_ex}
\end{figure}

To do so, we keep track of the \textit{exploration trace} $E_i$ of each robot $R_i$, defined as an area around the trajectory it has followed up to the current exploration time $t$. This area is calculated by integrating the robot's footprint across the trajectory. 
The robot's footprint is assumed to be a circle centered at the robot's location and with a radius equal to the communication range $d$.
An example of an execution trace computed with this method is shown in Fig.~\ref{fig:blob_ex}.

\begin{figure}
    \centering
    \includegraphics[width=0.11\linewidth]{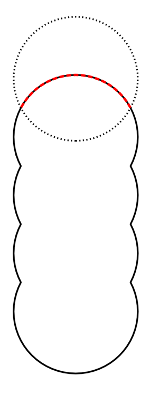}
    \includegraphics[width=0.63\linewidth]{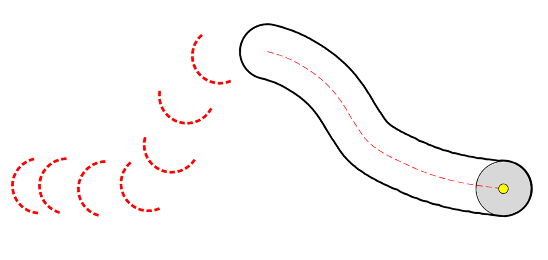}
    \caption{The oldest pose in the exploration trace is forgotten due to information decay (dashed black on the left). Frontiers in dashed red are created using our \textit{information decay} method.}
    \label{fig:information_decay}
    \vspace{-0.5cm}
\end{figure}

A key part of our method is to augment the set of frontiers $F$ with additional "virtual" frontiers that do not adhere to the standard definition but instead represent regions of interest for rendezvous. These frontiers, which we denote with the set $\bar{F}$, are obtained by using an information decay mechanism on the exploration trace. More precisely, at time $t$, we consider a portion of the exploration trace that is limited to the sub-trajectory linking the last $k$ poses. Denoting with $p^i_t$ the pose of robot $i$ at time step $t$, the trace is represented by the set $P^i_{k} = \{p^i_{t-k+1}, \ldots, p^i_t\}$. The trace portion starts at the robot's current position and are evenly spaced from there along the past visited locations. Each pose is characterized by a timestamp indicating when it was obtained; poses are removed after a time $T$ has passed after their acquisition, hence partitioning the trace into an active head and a vanishing tail. Each time a pose and the corresponding footprint on the trace are removed, a new virtual frontier $\bar{f} \in \bar{F}$ is created as the contour of the difference between the areas before and after the removal. A visual intuition of this process and its result is depicted in Fig.~\ref{fig:information_decay}.

As a result of this, while in a standard frontier-based exploration the robot selects the next most promising location from $F$, in our method the robot selects the most promising location from  $F \cup \bar{F}$. In this way, the robot is pushed to partially backtrack on its steps to revisit parts of the environments that have already been explored.

We rank frontiers (real and virtual) according to a linear combination of their distance from the current position of the robot, and their length, similar to what is proposed in~\cite{Yamauchi98}. Frontiers that are large and close to the robot are considered the most promising ones. More precisely, indicating with $dist(f)$ the distance between the current position of the robot and the frontier, and with $len(f)$ its length, the cost of a frontier is indicated as $\theta(f) = \alpha*len(f) - (1-\alpha) *dist(f)$. Note that, as our method adds new frontiers, it can be used with other exploration strategies.

\subsection{Online cluster formation}

We define a \textit{cluster} $C=\{R_i,R_j, \ldots \}$ as a set of one or more robots that can connect as per the communication model introduced above. Two clusters may merge to form a new one in an online fashion, by means of an episodic rendezvous as soon as at least one robot from each cluster comes within range of a counterpart from another cluster. When a new cluster is formed, one robot is elected as \textit{leader}, while the other members are termed \textit{followers}. (Since the robots form a connected network, this role assignment can be obtained collaboratively, by running a distributed consensus protocol.) The designated leader continues the frontier-based exploration process, with the remaining robots reacting to the leader's decisions by following it, keeping in formation, and trying not to be left out of the cluster. Clusters could split, for example when one or more robots get stuck in cluttered regions of the environment and lose connectivity with the rest of the group. In such instances, the isolated cluster (potentially a single robot) finding itself with no leader, elects a new one (possibly itself) and resumes the exploration process.


In determining the next location to reach for continuing the exploration, the leader uses a set of frontiers $F_C$ that is obtained by computing the boundaries between mapped and unknown spaces over a cluster-wise merged map, which integrates all the perceptions taken by each member of the newly formed cluster up to that time.

\begin{figure}
    \centering
    \includegraphics[width=1\linewidth]{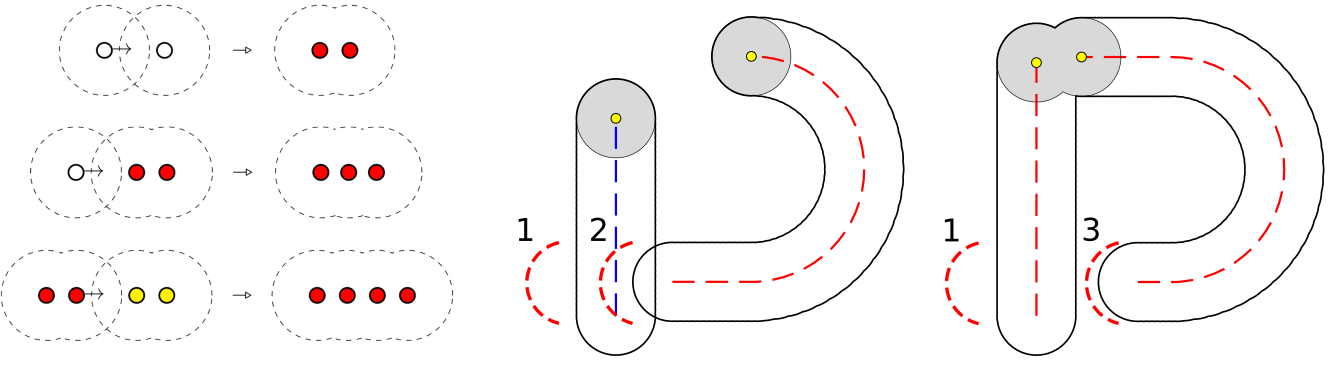}
    \caption{(left) Cluster formation; (right) Two robots meet and share their exploration trace. The frontiers are numbered. The red robot's frontier (2) is deleted as it overlaps with the exploration trace of the blue robot.}
    \label{fig:blob_sharing}
\end{figure}

The method based on information decay, previously introduced for generating additional virtual frontiers for a single robot, is now extended to clusters of two or more robots. Specifically, exploration traces are merged and virtual frontiers are computed from the decayed portions of each robot's trace. Then, any portion of frontiers falling within the still-not-vanished trace of any other robot in the cluster is deleted. This final step is performed to guarantee that all virtual frontiers are located in areas forgotten by the entire cluster, meaning that no robot has recently surveyed these regions. Consequently, these areas become potential rendezvous points for the entire cluster. Notice how this process can result in frontiers being divided into two or more separate parts. Analogously to the single-robot case, this set of additional frontiers, called $\bar{F}_C$ is included in the frontier-based exploration run by the leader which, iteratively, will select and communicate to the cluster the next location to reach considering $F_C \cup \bar{F}_C$. An example of this process is shown in Fig.~\ref{fig:blob_sharing}.



After a cluster is formed, only the leader of the cluster adds frontiers to $F_C$ and expands its exploration trace. 
Clearly, the rendezvous is considered accomplished when all robots belong to the same cluster.


\section{Experimental Results}

\begin{figure*}[t]
    \centering
    \includegraphics[trim={0cm 0cm 0cm 0cm},clip,width=\linewidth]{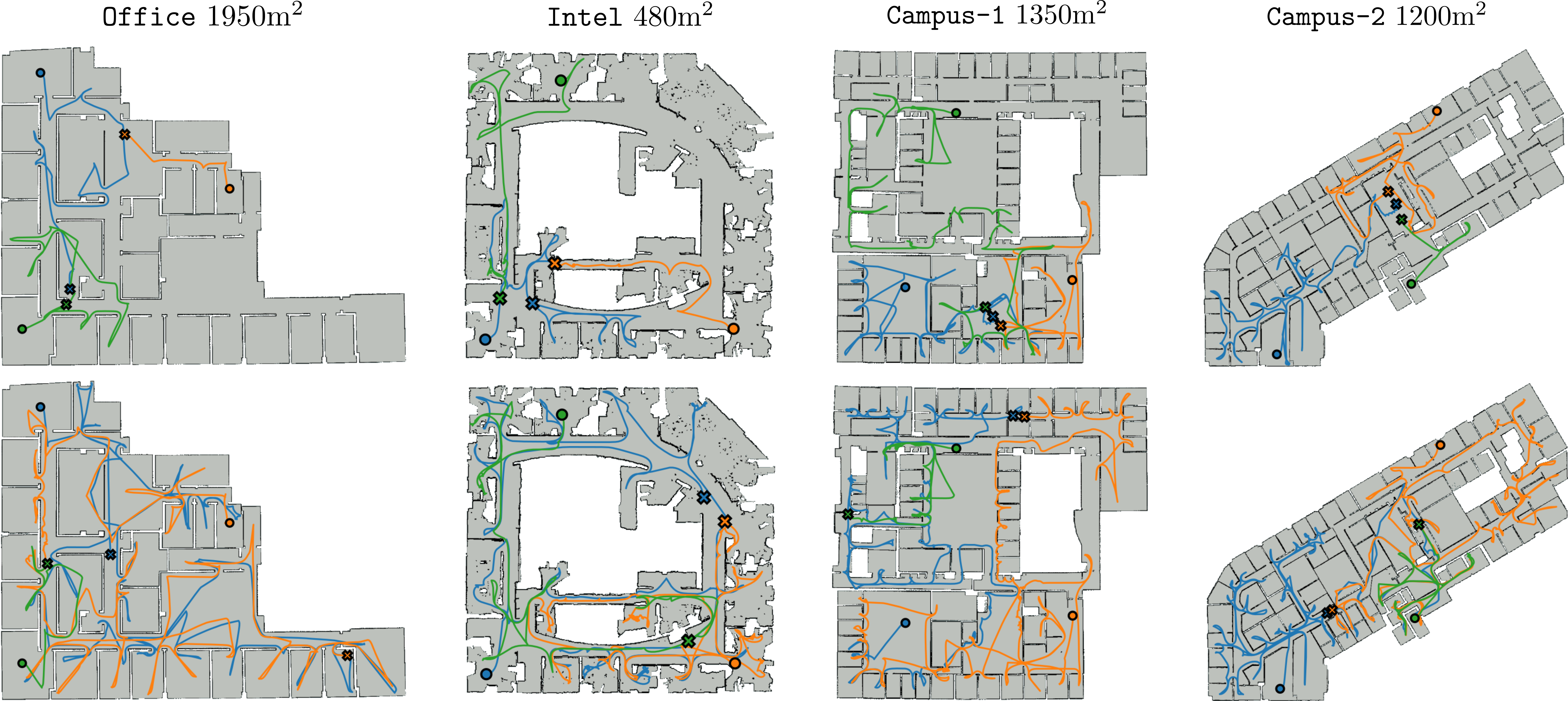}
    \caption{Rendezvous paths obtained with our method (\texttt{FBR} -- first row) and the baseline (\texttt{FBE} -- second row) in a series of environments. }
    \label{fig:maps}
    \vspace{-0.55cm}
\end{figure*}

\subsection{Experimental Setting}

We evaluate our method by implementing it in a real multi-robot control setup based on ROS. Thanks to the \emph{namespace} system provided by ROS, we can concurrently run different and logically separated control stacks, one for each robot, inside the same ROS environment. With such a control setup, we perform runs on environments simulated with Gazebo. We used Turtlebot3-burger robots with a speed of \SI{0.3}{\meter \per \second} and with a 2D lidar with a range of \SI{10}{\meter}; each robot performs SLAM using Gmapping~\cite{gmapping2007tro}, and runs a navigation stack with \href{http://wiki.ros.org/move_base}{\textcolor{blue}{move base}}. We simulate sensors and actuation errors, to pose challenges comparable to those of a real-world run. After preliminary experiments, we set information decay so that poses are forgotten each $\approx \SI{5}{\minute}$  and a communication range of $d=\SI{2.7}{\meter}$. A new pose is added to a trace each $\SI{2}{\second}$, while a frontier is added to $\bar{F}$ after 9 poses have been collected ($\SI{18}{\second}$). For the frontier evaluation function, we set $\alpha = 1/4$. Initially, we implemented leader-follower dynamics with the method from~\cite{testa2021choirbot} incurring in a significant number of robot failures (e.g., when trying to follow the leader through a narrow passage). Since the development of a robust formation-control mechanism is outside the scope of this paper, in our experiments we assumed perfect formation-control dynamics. With this setup, we compare our method (label \texttt{FBR}) against a baseline running a standard frontier-based exploration as in~\cite{Yamauchi98} (label \texttt{FBE}).



For each method and team of robots, we performed $10$ runs with random initial locations in $5$ indoor environments, shown in Figs.~\ref{fig:path_example} and~\ref{fig:maps}. \texttt{Ring} and \texttt{Office} are large-scale environments with $\approx40$ rooms we used to assess performance with a different number of robots, specifically from $3$ to $8$. We then evaluated, for teams of 3 robots, particularly challenging situations. First, we use the \texttt{Intel} environment~\cite{radish}, which is cluttered and with a less regular profile of obstacles. Then, we use \texttt{Campus-1} and \texttt{Campus-2}~\cite{IROS18}, two large-scale buildings with $\approx100$ rooms each.



Overall, we performed more than $180$ real-time exploration runs, recording the time $t$ to perform a rendezvous among all robots, the progression with time of $\max \left\vert C \right\vert$ (the size of the largest cluster), and the combined area explored by the robots $A$ obtained by computing the intersection of all maps aligned to the same reference system. 
A run is ended at time $t$ when $\max \left\vert C \right\vert$ is equal to $m$, or when all robots explore the entire environment (no frontiers left). This last case can occur only with the baseline since our method, by definition, keeps adding virtual frontiers. To register a performance in this degenerate case, we assume that robots travel to a default fallback location at the center of the map. The time to reach such a location is not included in the total.

We indicate as success rate $R$ as the percentage of runs concluded with a rendezvous, and the time $t_i$ as the time required for $i \in \{1,\ldots, N$\} robots to do a rendezvous. Our average results do not include runs where one or more robots experienced a permanent failure during navigation (e.g., due to a collision or because incapable of maneuvering away from a too-close obstacle). In our realistic setup, these kinds of events are fairly frequent, but in the majority of cases, robots managed to resolve the impasse by triggering a recovery behavior on the navigation stack.





In the following, we provide a significant selection of the obtained results. The used code and the full set of experimental results are available in a public repository\footnote{\scriptsize{\url{https://aislab.di.unimi.it/research/fberendezvous/}}}.

\subsection{Results}
\begin{figure}[h]
    \vspace{-0.2cm}
    \centering
    \includegraphics[width=\linewidth]{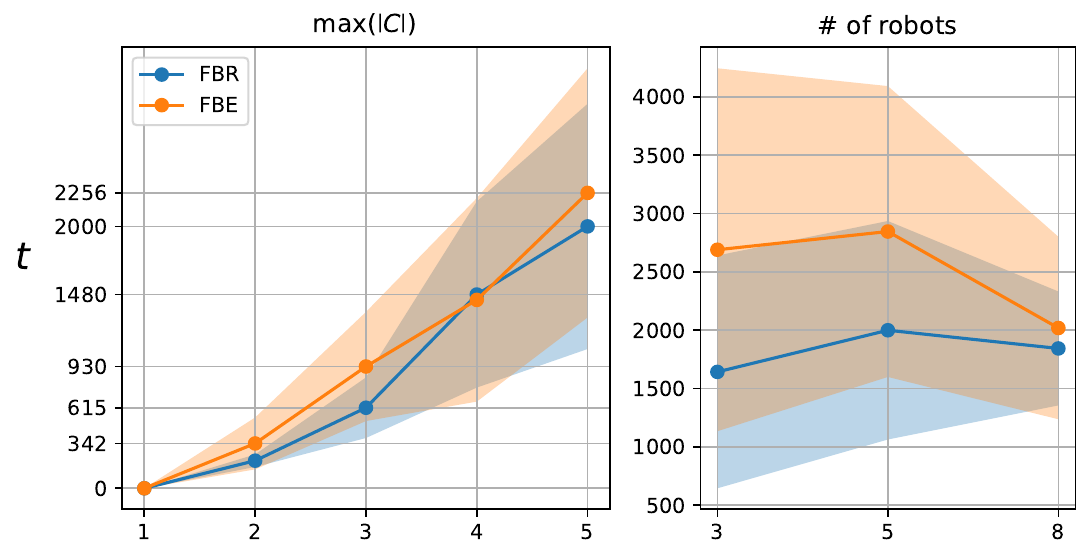}
    \caption{(left) The average time $t$ (and standard deviation) to complete the rendezvous as the maximum size of the cluster changes in the environment \texttt{ring} when an MRS of $5$ robots is used; note how, in this environment the $R=0.7$ for \texttt{FBE}.
    (right) The average time $t$ (and standard deviation) required for a run in \texttt{ring} when a different number of robots is used.}
    \vspace{-0.2cm}
    \label{fig:mean_times}
\end{figure}

The results are reported in Table \ref{tab:RESULTS}.
We first discuss the results obtained in \texttt{Ring} and \texttt{Office}, using 3, 5, and 8 robots.
The first interesting remark that emerges is that a frontier-based exploration strategy can be used, with fair results, to perform a rendezvous in an initially unknown environment. At the same time, the use of our method significantly improves the chances of episodic rendezvous, as can be seen by the gain in terms of $\Delta_t = (t_{\texttt{FBR}}-t_{\texttt{FBE}})/t_\texttt{FBE}$, where $t_{\texttt{FBR}}$ and $t_{\texttt{FBE}}$ are the times required by \texttt{FBR} and \texttt{FBE}, respectively. This can be seen from the different paths resulting from the two methods, as shown in Fig. \ref{fig:path_example} and Fig. \ref{fig:maps}.
The improvement is more evident when the number of robots is small, and the environment is large, i.e., when there are a few robots distributed across a large map.
While our method allowed the whole team of robots to perform a rendezvous in all of the $60$ runs of this type, with $R=1$, there is a non-negligible chance that robots are not able to perform a rendezvous if a standard frontier-based exploration method \texttt{FBE} is used. This usually happens when, by chance, robots in the early stages of exploration choose different directions; 
during the remainder of the exploration run, the robots switched their location without meeting the other members of the MRS. An example of this can be seen in Fig. \ref{fig:path_example}(right). 
The use of our method, which encourages the robot to backtrack on previously explored positions, is effective in facilitating rendezvous. At the same time, the performance is far more stable than those of \texttt{FBE}, with a lower standard deviation.  These findings are reported in Fig. \ref{fig:mean_times}, which shows the result in terms of time $t$ for 3 to 8 robots in \texttt{Ring}.
When more robots are used, the chances of having episodic rendezvous are higher, and thus the differences between the two methods are less evident. In that case, the standard \texttt{FBE} can be effectively used also to perform rendezvous.

\renewcommand{\arraystretch}{1.1}
\begin{table}[t]
\begin{adjustbox}{width=\columnwidth,center}
\begin{tabular}{c|c|c|c|c|c|c|c|c}
\cline{3-8}
\multicolumn{1}{l}{}    & \multicolumn{1}{l|}{} & \multicolumn{3}{c|}{\texttt{FBR}}          & \multicolumn{3}{c|}{\texttt{FBE}}            & \multicolumn{1}{l}{} \\ \hline
\multicolumn{1}{|c|}{\textbf{Environment}}       & \textbf{$m$}          & $t$ (s) & $\sigma_t$ & $R$ & $t$ (s) & $\sigma_t$ & $R$ & \multicolumn{1}{c|}{$\Delta_t$}    \\ \hline
\multicolumn{1}{|c|}{\multirow{3}{*}{\texttt{Ring}}}   & 3                    & 1623.13        & 978.18       & 1          & 2691.64        & 1557.18      & 0.6        & \multicolumn{1}{c|}{0.66}                 \\ \multicolumn{1}{|c|}{}
                        & 5                    & 2000           & 936.3        & 1          & 2846.06        & 1246.36      & 0.7        & \multicolumn{1}{c|}{0.42}                 \\ \multicolumn{1}{|c|}{}
                        & 8                    & 1844.08        & 488.16       & 1          & 2003.4         & 782.05       & 1          & \multicolumn{1}{c|}{0.09}                 \\ \hline
\multicolumn{1}{|c|}{\multirow{3}{*}{\texttt{Office}}} & 3                    & 727.46         & 372.16       & 1          & 851.11         & 791.46       & 0.9        & \multicolumn{1}{c|}{0.17}                 \\ \multicolumn{1}{|c|}{}
                        & 5                    & 1358.55        & 718.19       & 1          & 1478.21        & 594.59       & 8.8        & \multicolumn{1}{c|}{0.02}                 \\ \multicolumn{1}{|c|}{}
                        & 8                    & 1309.45        & 437.5        & 1          & 1251.54        & 762.47       & 1          & \multicolumn{1}{c|}{-0.04}                \\ \hline
\multicolumn{1}{|c|}{\texttt{Intel}}                   & 3                    & 580.13         & 256.63       & 1          & 688.19         & 553.28       & 0.9        & \multicolumn{1}{c|}{0.19}                 \\ \hline
\multicolumn{1}{|c|}{\texttt{Campus-1}}                 & 3                    & 1070.29        & 487.82       & 1          & 1591.7         & 1130.81      & 1          & \multicolumn{1}{c|}{0.49}                 \\ \hline
\multicolumn{1}{|c|}{\texttt{Campus-2}}                 & 3                    & 818.09         & 503.57       & 1          & 1370.68        & 959.97       & 1          & \multicolumn{1}{c|}{0.68}     \\ \hline           
\end{tabular}
\end{adjustbox}
\caption{
Results of \texttt{FBR} compared with \texttt{FBE} in all of the environments $t$ is in seconds, $R$ is in $[0,1]$.}\label{tab:RESULTS}
\vspace{-0.5cm}
\end{table}

We then compare the results obtained in the three complexes \texttt{Intel}, \texttt{Campus-1}, \texttt{Campus-2}, environments using $3$ robots.  
While the three environments present different types of challenges, the results obtained confirm our findings and the robustness of our approach. On one side, our method is always able to conclude the run with a rendezvous ($R=1$), while requiring less time when compared with \texttt{FBE}. 

Interestingly, the areas of the maps acquired using our approach are close to the ones acquired using \texttt{FBE}. On average, a team of robots using our method performs a rendezvous while exploring  $3.3\% (\sigma=7.7)$ less area than when using \texttt{FBE} in the same condition. The results are stable across all $5$ environments and with different numbers of robots and are omitted for brevity. Consequently, the robots using our method not only meet faster but also acquire the same amount of knowledge about the working environment while performing exploration to do a rendezvous. 

\section{Conclusion and future works}

In this work, we have presented a framework for allowing an MRS to perform a rendezvous in a previously unknown environment while performing exploration. To do so, we have introduced a mechanism for information decay on top of a frontier-based exploration approach. 
In this way, the robots are biased to backtrack on previously explored parts of the environment, thus facilitating accidental rendezvous in high-connectivity parts of the environment such as corridors and hallways.
The experimental evaluation clearly shows how our framework allows the robot to perform a rendezvous in less time than a classic frontier-based exploration approach. At the same time, we point out how standard frontier-based exploration methods can be used to perform rendezvous when the number of robots is higher. 

\bibliographystyle{IEEEtran}
\bibliography{citations}
\end{document}